\documentclass{article}

     \PassOptionsToPackage{numbers, compress}{natbib}

\usepackage[preprint]{neurips_2024}

\usepackage[utf8]{inputenc} 
\usepackage[T1]{fontenc}    
\usepackage{hyperref}       
\usepackage{url}            
\usepackage{booktabs}       
\usepackage{amsfonts}       
\usepackage{nicefrac}       
\usepackage{microtype}      
\usepackage{xcolor}         
\usepackage{graphicx}
\usepackage{multirow}
\usepackage{caption}
\usepackage{array}
\usepackage{tikz}
\usetikzlibrary{calc}
\usepackage{graphicx}
\usepackage{threeparttable,booktabs}
\usepackage{wrapfig}
\usepackage{soul}           
\usepackage{xcolor}
\usepackage{ragged2e}
\newcolumntype{P}[1]{>{\RaggedRight\hspace{0pt}}p{#1}}
\usepackage{svg}

\def\DEBUG{} 

	\ifdefined\DEBUG
	    \newcommand{\gaurav}[1]{{\color{blue}[Gaurav: #1]}}
	    \newcommand{\akash}[1]{{\color{cyan}[Akash: #1]}}
	    \newcommand{\ravisri}[1]{{\color{red}[Ravisri: #1]}}
        \newcommand{\kushal}[1]{{\textcolor{brown}[Kushal: #1]}}
	\else
	    \newcommand{\gaurav}[1]{}
	    \newcommand{\akash}[1]{}
	    \newcommand{\ravisri}[1]{}
        \newcommand{\kushal}[1]{}
	\fi
\title{Scaling the Vocabulary of Non-autoregressive Models for Efficient Generative Retrieval}

\author{%
\textbf{Ravisri Valluri$^1$} \quad \textbf{Akash Kumar Mohankumar$^2$} \quad \textbf{Kushal Dave$^3$}\\ \quad \textbf{Amit Singh$^2$} \quad \textbf{Jian Jiao$^3$} \quad \textbf{Manik Varma$^1$} \quad \textbf{Gaurav Sinha$^1$} \\
$^1$Microsoft Research, India \quad $^2$Microsoft, India \quad $^3$Microsoft, USA\\
\texttt{\{t-ravalluri, makashkumar, kudave\}@microsoft.com}\\
\texttt{\{siamit, jian.jiao, manik, gauravsinha\}@microsoft.com}
}

\begin{document}

\maketitle

\begin{abstract}
Generative Retrieval introduces a new approach to Information Retrieval by reframing it as a constrained generation task, leveraging recent advancements in Autoregressive (AR) language models. However, AR-based Generative Retrieval methods suffer from high inference latency and cost compared to traditional dense retrieval techniques, limiting their practical applicability. This paper investigates fully Non-autoregressive (NAR) language models as a more efficient alternative for generative retrieval. While standard NAR models alleviate latency and cost concerns, they exhibit a significant drop in retrieval performance (compared to AR models) due to their inability to capture dependencies between target tokens. To address this, we question the conventional choice of limiting the target token space to solely words or sub-words. We propose PIXAR, a novel approach that expands the target vocabulary of NAR models to include multi-word entities and common phrases (up to 5 million tokens), thereby reducing token dependencies. PIXAR employs inference optimization strategies to maintain low inference latency despite the significantly larger vocabulary. Our results demonstrate that PIXAR achieves a relative improvement of 31.0\% in MRR@10 on MS MARCO and 23.2\% in Hits@5 on Natural Questions compared to standard NAR models with similar latency and cost. Furthermore, online A/B experiments on a large commercial search engine show that PIXAR increases ad clicks by 5.08\% and revenue by 4.02\%.

\end{abstract}
\section{Introduction}
\label{section:intro}
Generative Retrieval (GR) has emerged as a promising approach within Information Retrieval, particularly for text retrieval tasks \cite{SEAL, DSI, NCI, MINDER}. This approach involves creating a set of document identifiers that represent documents from the original corpus. A generative model is then trained to generate document identifiers for an input query. The generated identifiers are subsequently mapped back to the corresponding documents in the corpus. GR methods typically utilize an autoregressive (AR) language model to generate the document identifier as a sequence of words or sub-words tokens from a predefined target vocabulary. By leveraging high-quality document identifiers and capturing complex dependencies between tokens through the autoregressive generation process, GR has achieved substantial improvements in retrieval performance in recent years \cite{SEAL, MINDER, LTRGR}.

Despite these advancements, deploying GR models in low-latency applications, such as sponsored search, remains a significant challenge due to the high inference complexity of AR models \cite{generative_retrieval_survey, unity}. This stems from their sequential token-by-token generation mechanism \cite{nar_translation}. To address this challenge, our paper explores the use of non-autoregressive (NAR) language models for GR. These models significantly reduce inference costs by generating all tokens of the document identifier simultaneously. However, this parallel generation limits the model's ability to capture dependencies among tokens (words, sub-words) in the output identifier, leading to inferior retrieval performance compared to AR-based GR models. To enable NAR-based GR to leverage word and sub-word interactions during generation, we propose expanding the model's target vocabulary by incorporating phrases within the document identifiers as tokens. Intuitively, predicting high-probability phrases at each position in the output sequence allows the NAR model to better understand the intricate relationships between words and sub-words within each predicted phrase, potentially enhancing retrieval performance. This forms the basis of our first research question:

\textbf{(RQ1)-} How does the retrieval accuracy of a NAR-based GR model (with a target vocabulary containing word/sub-word level tokens) change when the target vocabulary is expanded to include phrases from document identifiers as additional tokens? While a positive answer to the above question will provide an approach to get high quality retrieval from NAR-based GR, it also comes at a cost to the inference latency. While generating phrases at output instead of solely words leads to shorter output sequences and helps latency, as the vocabulary size grows, predicting the most likely tokens at each of these output positions becomes computationally far more demanding leading to much higher overall latency. Consequently, to make NAR-based GR truly viable for latency-sensitive applications, we need to develop efficient inference methods that can select the top tokens from the enlarged vocabulary more efficiently. This leads us to our second research question:

\textbf{(RQ2)-} How can we reduce the inference latency of a NAR-based GR model with a large target vocabulary without compromising its retrieval accuracy?

In this work, we make progress on both these questions. Our key contributions are outlined below.

\subsection{Our Contributions}
\begin{enumerate}
    \item We present PIXAR (Phrase-Indexed eXtreme vocabulary for non-Autoregressive Retrieval), a novel approach to NAR-based GR. By leveraging a vast target vocabulary encompassing phrases within document identifiers, PIXAR achieves superior retrieval quality compared to conventional NAR-based GR models. Through innovative training and inference optimizations, PIXAR effectively mitigates the computational burden associated with its large vocabulary. This allows for efficient retrieval of relevant documents during the inference process. The architecture of PIXAR is presented in Figure \ref{fig:pixar}.  A comprehensive explanation of each component can be found in Section \ref{section:proposed-work}.
    \item We conducted comprehensive experiments on two widely-used text retrieval benchmarks, MS MARCO \cite{bajaj2018ms} and Natural Questions (NQ) \cite{NQ}. Our results demonstrate PIXAR's significant performance gains: a relative improvement of 24.0\% in MRR@10 on MSMARCO and a 23.2\% increase in Hits@5 on NQ, compared to standard NAR-based retrieval models while maintaining similar inference latency. These findings underscore PIXAR's effectiveness in enhancing retrieval quality for various text retrieval tasks.  
    \item Moreover, A/B testing on a large commercial search engine revealed a significant impact of PIXAR: a 5.08\% increase in ad clicks and a 4.02\% boost in revenue. These findings validate PIXAR's practical value in improving user engagement and driving business outcomes.
\end{enumerate}

\section{Related Work}
\textbf{Generative retrieval:} GR is an emerging paradigm in information retrieval that formulates retrieval as a generation task. A key distinction among different GR methods lies in their approach to represent documents. Some methods directly generate the full text of the document, particularly for short documents like keywords \cite{endtoendBaidu, CLOVERv1, prophetnet_ads}. Others opt for more concise representations, such as numeric IDs \cite{DSI, DSI-QG, DSI++, NCI, RIPOR, recgen}, document titles \cite{GENRE, Chen2022CorpusBrainPA}, sub-strings \cite{SEAL}, pseudo queries \cite{Tang2023SemanticEnhancedDS}, or a combination of these descriptors \cite{MINDER, LTRGR}. Despite showcasing promising results, existing GR approaches have high inference latency and computational cost due their reliance on AR language models, presenting a significant challenge for their real-world adoption. 

\textbf{Non-autoregressive Models:} Recent works have explored NAR models for various generation tasks, such as machine translation \cite{nar_translation}, text summarization \cite{bang}, and specific retrieval applications like sponsored search \cite{unity}. NAR models aim to accelerate inference by predicting word or sub-word tokens independently and in parallel with a single forward pass. However, NAR models struggle to capture the inherent multimodality in target sequences, where multiple valid outputs exist for a single input, due to their lack of target dependency modeling \cite{nar_translation}. This often leads to predictions that mix tokens from multiple valid outputs, resulting in significant performance degradation. To mitigate this, existing approaches focus on accurately predicting a single mode rather than modeling all modes. For instance, some methods use knowledge distillation to simplify the training data \cite{nar_translation, understanding_kd}, while a few others relax the loss function \cite{axe, oxe, ctc_1, ctc_2}. While these approaches are effective for tasks requiring a \textit{single} correct output, GR necessitates retrieving \textit{all} relevant document identifiers for accurate retrieval and ranking. In this work, we propose an orthogonal approach to improve NAR models for retrieval by directly predicting phrases instead of sub-words. This reduces the number of independent predictions required in NARs, leading to improved retrieval performance.

\textbf{Efficient Softmax:} The softmax operation, crucial for generating probability distributions over target vocabularies in language models, presents a significant computational bottleneck, particularly for large vocabularies. Existing approaches address this through techniques such as low-rank approximation of classifier weights \cite{fastsoftmax, hire}, clustering of classifier weights or hidden states to pre-select target tokens \cite{adaptivesoftmax, l2s}. However, these methods remain computationally expensive for NAR models which perform multiple softmax operations within a single forward pass. In contrast, we introduce a novel method that utilizes a dedicated shortlist embedding to efficiently narrow down target tokens for the entire query, thereby significantly reducing latency and maintaining strong retrieval performance.

\textbf{Large Vocabulary:} Recent work has highlighted the benefits of large sub-word vocabularies for encoder models, particularly in multilingual settings \cite{xlmv}. Non-parametric language models, which predict outputs from an open vocabulary of n-grams and phrases using their dense embeddings, have also gained traction for tasks like question answering and text continuation \cite{npm_lm, retrieval_is_acc_gen, cog}. While our work shares the goal of expanding vocabulary size with non-parametric models, we directly learn classifier weights for an extended target vocabulary within a non-autoregressive framework. 

\section{Preliminaries}
\label{section:preliminaries}
\textbf{Notation:} We let $\mathcal{Q}$ to be a set of queries and $\mathcal{X}$ to be a finite set of textual documents (called the document corpus). Following the GR paradigm from prior works \citep{SEAL, MINDER, LTRGR}, we use a set of document identifiers (docids) $\mathcal{D}$. Prior literature uses docids such as titles, sub-strings, pseudo-queries etc. In this paper, following recent works \cite{NCI, DSI-QG, MINDER}, we leverage pre-trained language models to generate high quality pseudo-queries from documents, which we then use as docids. For non-negative integers $m<n$, we denote the set $\{m, \ldots, n\}$ by $[m,n]$. We use $\mathbb{P}$ (with or without subscripts) to denote probability distributions and the exact distribution is made clear at the time of use. Next we describe salient features of NAR language models relevant to our work.


\textbf{NAR Models:} NAR models generate all tokens of the docid in parallel and therefore lead to faster retrieval than AR models. These models assume conditional independence among target tokens, i.e., $
\mathbb{P}(d \mid q, \theta) = \prod_{t = 1}^{n} \mathbb{P}(d^t \mid q, \theta)$ and so for each position $t\in [s]$, they select the top tokens based on the conditional probability distribution $\mathbb{P}(. \mid q, \theta)$. This simplification enables efficient inference but comes at a cost.  Previous studies in various applications, including machine translation \cite{nar_translation, nar_tricks_of_trade}, have demonstrated that the assumption of conditional independence rarely holds for real-world data. Consequently, NAR models often struggle to capture crucial dependencies between target tokens, leading to a substantial performance degradation compared to their autoregressive counterparts. In our proposed work described in Section \ref{section:proposed-work}, we develop a technique that can overcome this quality degradation by adding phrase level tokens (within docids) and designing novel training/inference mechanisms that can still benefit from the parallel generation mode of NAR models.

\section{Proposed Work: PIXAR}
\label{section:proposed-work}
\begin{figure}
    \centering
\centering
\includegraphics[width=0.85\textwidth]{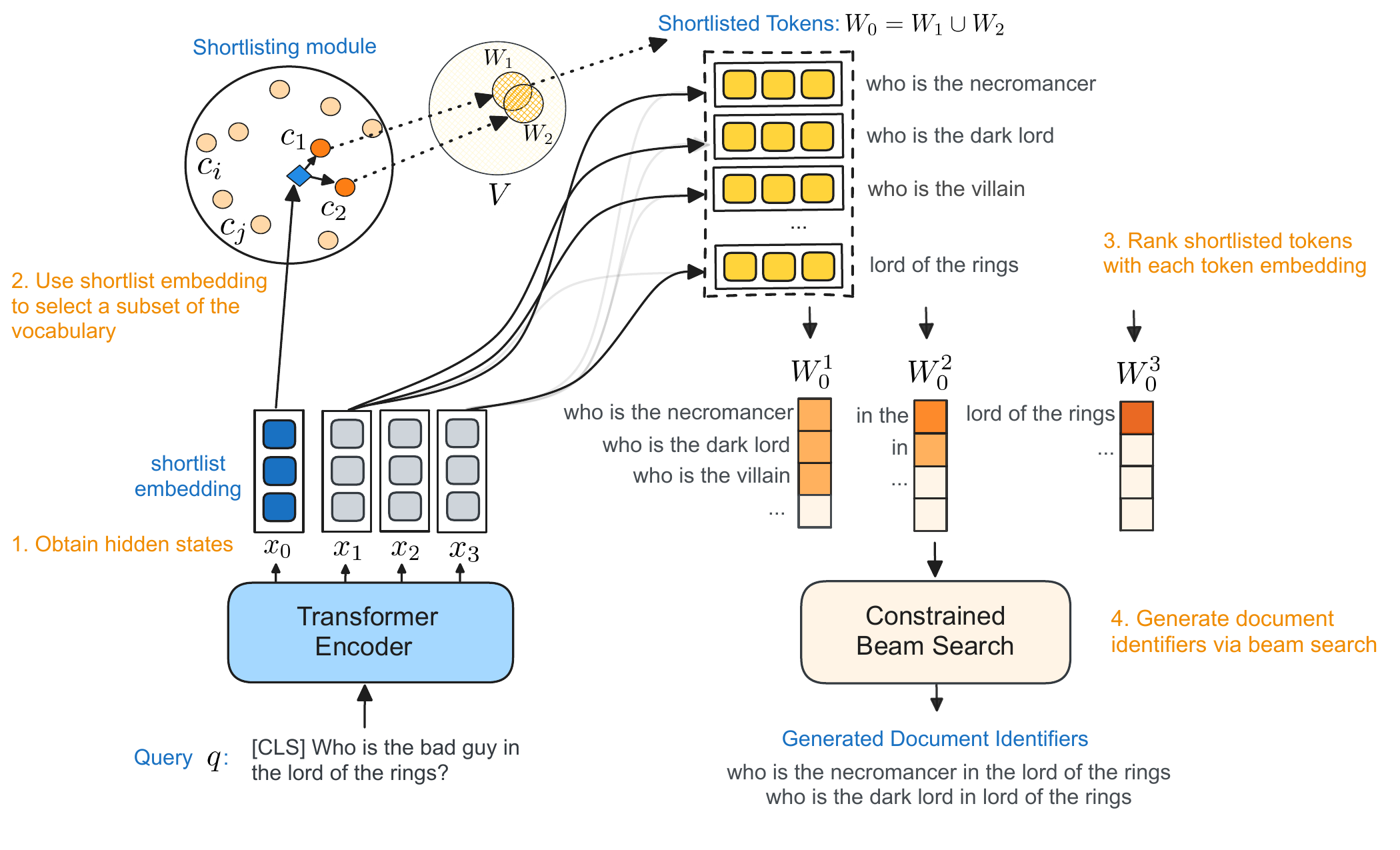}
    \caption{PIXAR inference pipeline: The query is first encoded by a Transformer to produce shortlist embedding $x_0$ and token embeddings \{$x_1, \cdots, x_s$\}. The shortlist embedding $x_0$ is used to identify $k$ vocabulary clusters $\{c_i\}_{i=1}^{k}$. The union of these clusters, $W_0$, is then re-ranked at each position using the corresponding token embeddings, producing a set of ranked candidate tokens ($W_0^1$ ... $W_0^s$) with their probability scores. The docids are predicted from these tokens via constrained beam search}
    \label{fig:pixar}
\end{figure}

The core idea behind PIXAR is to scale up the target vocabulary of NAR models by including phrases from docids. We explain the methodology for constructing this expanded vocabulary in Section \ref{subsec:vocab_selection}. To enable efficient inference with a larger vocabulary, PIXAR constructs a small number of token subsets from the target vocabulary during training. At inference time, PIXAR selects and combines relevant subsets to create a concise \emph{shortlist} of candidate tokens. For each output position,  PIXAR only re-ranks tokens among this shortlisted subset to predict the top tokens. Finally, these top tokens at different positions are combined using trie constrained beam search to generate the docids. Section \ref{subsection:pixar-pipeline} provides the complete details of the PIXAR pipeline, including the novel training and inference mechanisms. Figure \ref{fig:pixar} illustrates the different components of PIXAR through a concrete example.

\subsection{Vocabulary Construction}
\label{subsec:vocab_selection}
Our goal is to build a target tokenizer and vocabulary with the following desired characteristics: (i) Efficient Encoding: The vocabulary should encode docids using fewer bits, resulting in shorter target sequences, (ii) Token Frequency: Ensure every token appears with a minimum frequency in the docid set to facilitate effective training of the language modelling weights, and (iii) Linguistic Structure: Include common phrases while respecting word boundaries. While Byte-Pair Encoding (BPE) \cite{BPE} is a popular method for constructing vocabularies, its greedy merging strategy often results in sub-optimal tokens that blend characters from different words. To circumvent this, tokenizers in language models like LLama \cite{llama2} and GPT \cite{gpt2, gpt4} incorporate a pre-tokenization heuristic that splits sentences by spaces and punctuation marks before applying BPE. However, this approach results in vocabularies limited to words and sub-words, which, as we show in Section \ref{subsec:ablation}, perform significantly worse than phrase-based vocabularies. 

Instead, we adopt a two-stage approach: candidate selection followed by vocabulary construction, as proposed in TokenMonster \cite{tokenmonster}. Initially, we generate a set of potential token candidates by considering all possible character substrings up to a specified maximum length. We then filter these substrings based on criteria such as adherence to word boundaries and consistency in character types (letters, numbers, punctuation, etc.). Only tokens that exceed a minimum occurrence threshold are retained as potential candidates. In the second stage, we iteratively refine the candidate set to construct an optimal vocabulary of a specified size. We generate multiple test vocabularies from the pool of candidate tokens. Each test vocabulary is then used to tokenize the dataset, and a scoring system evaluates the efficiency of each token based on the total number of characters it compresses in the docid set. Tokens that perform poorly are removed, and the process is repeated until the desired vocabulary size is reached. Since we follow the vocabulary construction process from TokenMonster \cite{tokenmonster}, we refer the reader to \cite{tokenmonster} for further details.


\subsection{PIXAR Pipeline}
\label{subsection:pixar-pipeline}
In this section, we provide details of the PIXAR pipeline. 
At a high level PIXAR comprises of a NAR model, a set of learnable vectors $c_1, \ldots, c_m$ and their corresponding $r$-sized subsets $W_1,\ldots, W_m\subset V$, where $V$ is the target vocabulary constructed using the method described in Section \ref{subsec:vocab_selection}. Here, $m$ and $r$ are hyper-parameters that can be tuned. The set $W_i, i\in [m]$ contains the top $r$ tokens in the target vocabulary $V$ as per the Softmax probability score,
\[
\mathbb{P}_{c_i}(v) = \frac{\exp\left(c_i^T w_{v}\right)}{\sum\limits_{u\in V}\exp\left(c_i^T w_u\right)},
\]
where for each token $u\in V$, $w_u\in \mathbb{R}^d$ is a learnable parameter vector in the NAR model. We will explain the role of the $c_i$s below but first we demonstrate the journey of an input query $q$ through the pipeline. $q$ is first prepended with a special "[CLS]" token and sent through the NAR model. It passes through the transformer layers which outputs a sequence of embeddings $x_0(q), x_1(q), \ldots, x_s(q) \in \mathbb{R}^d$, where $s$ is the output sequence length and $d$ is the hidden dimension of the embeddings. Following this, $k$ vectors from the set $\{c_i, i\in [m]\}$, that have the largest inner product with $x_0(q)$ are computed. Without loss of generality, assume they are $c_1, \ldots, c_k$. The union of the corresponding sets i.e., $W_0(q) = W_1\cup\ldots\cup W_k$ is then obtained. This becomes a final set of shortlisted tokens from $V$ and tokens within it are subsequently used for generation of the docids. This means $W_0(q)$ should at least contain tokens for all positions in the output to be generated for $q$. For each $t\in [1,s]$, the set $W_0(q)$ is re-ranked according to the Softmax probability scores\footnote{we actually use estimates $\widetilde{P}_t(.\mid q)$ (described later) that can be computed efficiently.} $\mathbb{P}_t(.\mid q)$, defined as,
\[
\mathbb{P}_t(v\mid q) = \frac{\exp\left(x_t(q)^Tw_v\right)}{\sum\limits_{u\in V}\exp\left(x_t(q)^Tw_u\right)} 
\]
 This gives ordered sets $W_0^t(q)$ for each $t\in [1,s]$. The top tokens in $W_0^t(q)$ are ideally more relevant to the $t^{th}$ position in the docid to be generated. We generate the top docids by performing permutation decoding \cite{CLOVERv1} which utilizes constrained beam search on trie data structures representing document identifiers in $\mathcal{D}$ as a sequence of tokens from the target vocabulary $V$. Since $x_0(q)$ is used to obtain the shortlisted set of tokens $W_0(q)$, we call it the \emph{shortlist embedding}.

\textbf{Training:} We train PIXAR using a training dataset of query, docid pairs $(q_1, d_1), \ldots ,(q_N, d_N)$. Our training has two parts. First, we minimize a novel loss function $\ell(\bar\theta)$ to learn a vector $\bar\theta$ comprising of the the hidden parameters within the transformer layers as well as the token parameter vectors $w_u, u\in V$. Our loss $\ell(\bar\theta)$ comprises of three terms. 
The first term $\ell_1(\bar\theta)$ is the standard cross entropy loss between the Softmax predictions at each $t\in [1, s]$ and the actual docid sequence of the document identifiers in the training data, i.e.,
\[
\ell_1(\bar{\theta}) = -\sum\limits_{i=1}^N\sum\limits_{t=1}^s\log\bigg[\mathbb{P}_t(d_i^t | q_i)\bigg]
\]
In the PIXAR pipeline, we use $x_0(q)$ to compute a subset of tokens $W_0(q)\subset V$, that should ideally contain tokens at all positions in the generated output docid. To achieve this we add another cross entropy loss term that intuitively accounts for how well a Softmax activation is able to predict the set of tokens in the output docid by using embedding $x_0(q)$, i.e.,
\[
\ell_2(\bar{\theta}) = -\sum\limits_{i=1}^N\sum\limits_{t=1}^s \log\bigg[\mathbb{P}_0(d_i^t | q_i)\bigg]
\]

Finally, for all $t\in [1,s]$ we add a self normalization loss term that enables efficient computation of the Softmax based probability scores, i.e.,

\[
\ell_3(\bar{\theta}) = \sum\limits_{i=1}^N\sum\limits_{t=1}^s  \log^2\bigg[\sum\limits_{v\in V}\exp\left(x_t(q_i)^T w_v\right)\bigg]
\]
Note that, post minimization of the loss $\ell_3(\bar\theta)$, for each $t\in [1,s]$, we can use the probability estimates $\widetilde{\mathbb{P}}_t(v\mid q) = \exp\left(x_t(q)^Tw_v\right)$,
instead of $\mathbb{P}_t(v\mid q)$ defined above. For large target vocabularies (e.g., our expanded vocabulary with phrase tokens), these estimates are much faster to compute since the sum in the denominator over the entire target vocabulary is avoided. Finally, we combine these three terms into our overall loss as,
\[
\ell(\bar\theta) = \ell_1(\bar\theta) + \lambda_2 \ell_2(\bar\theta) + \lambda_3 \ell_3(\bar\theta),
\]
where $\lambda_2, \lambda_3$ are hyper-parameters to be tuned. After minimizing $\ell(\bar\theta)$, we train further to learn the vectors $c_1,\ldots, c_m$ described earlier. For each training pair $(q_i, d_i), i\in [N]$, let $e_i\in [m]$ be such that $c_{e_i}$ has the largest inner product with $x_0(q_i)$, i.e.,
\[
e_i = \arg\max_{j\in [m]}\langle x_0(q_i), c_j\rangle
\]

Then we minimize a function $\ell^\prime(c_1,\ldots,c_m)$ that computes the cross entropy loss between the Softmax distributions $\mathbb{P}_{c_{e_i}}, i\in [N]$ and the docid sequence $d_i$, i.e.,
\[
\ell^\prime(c_1,\ldots,c_m) = -\sum\limits_{i=1}^N\sum\limits_{t=1}^s \log\bigg[\mathbb{P}_{c_{e_i}}(d_i^t)\bigg] 
\]
Intuitively, this means that we try to maximize the likelihood of the tokens present in the docid $d_i$, for the vector $c_{e_i}$ that is most aligned with $x_0(q_i)$. This will ensure that the set $W_{e_i}$ (defined earlier in this section) will have a good chance of containing the tokens in $d_i$. Recall that, in our description of the PIXAR pipeline we find $k$ vectors that have highest inner product with $x_0(q_i)$ and not just the most aligned vector $c_{e_i}$. This enhances the chance of the tokens in $d_i$ being present in $W_0(q)$, since it is a union of the sets of tokens corresponding to these $k$ vectors.

\textbf{Efficient Inference:} We now explain how the PIXAR pipeline outlined in this section is able to circumvent latency overheads that arise due to the new expanded target vocabulary $V$. Recall the typical NAR model described in Section \ref{section:preliminaries}. As the size of $V$ becomes larger the computational cost of inference grows primarily due to two reasons; (a) the language modelling head needs to select top tokens from $V$ at each output position, and (b) computing the Softmax distribution at each output position becomes expensive since its denominator computes a sum over the target vocabulary. While (b) is easily tackled using the self normalization loss $\ell_3(\bar\theta)$, PIXAR's handling of (a) is more intricate. Instead of directly selecting tokens from $V$ at each output position, it selects tokens from re-ranked versions of the shortlisted subset $W_0(q)$. This set is further a union of ($k$ many) $r$-sized subsets and therefore has size $\leq rk$. By choosing hyper-parameters appropriately, we can ensure that $rk\ll |V|$. To identify the shortlisted subset $W_0(q)$, PIXAR finds the $k$ vectors among $c_1,\ldots,c_m$ with largest inner product with $x_0(q)$. Given $x_0(q)$, this can also be done efficiently since we can choose the hyper-prameter $m$ appropriately, i.e. $m\ll |V|$. This allows the PIXAR pipeline to avoid the additional inference latency that arises from the expanded target vocabulary $V$. Note that very small values of $m,r,k$ can impact retrieval quality and therefore need to be tuned for high quality retrieval. In our experiments in Section \ref{section:experiments}, we demonstrate for two popular datasets that even when $|V|$ is scaled to $5$ million, $m,r,k$ can be chosen in a way that ensures high retrieval quality with negligible impact on inference time.

\section{Experiments \& Results}
\label{section:experiments}
In this section, we evaluate our proposed PIXAR method in three different experimental settings. First, we benchmark PIXAR against leading GR approaches, including AR and NAR methods. Next, we perform a component-wise ablation study on PIXAR to examine the impact of each component on retrieval performance and model latency. We also compare our novel inference pipeline (Section \ref{subsection:pixar-pipeline}) with inference optimization methods from the literature. Finally, we assess the effectiveness of PIXAR in a real-world application, focusing on sponsored search.

\subsection{Experimental Setup}
We evaluate PIXAR on two types of datasets: (i) public datasets designed for passage retrieval tasks, and (ii) a proprietary dataset used for sponsored search applications. Below, we describe each dataset:

\textbf{Public Datasets:} We use two prominent datasets to evaluate PIXAR and other GR methods: MS MARCO \cite{bajaj2018ms} and Natural Questions (NQ) \cite{NQ}. The MS MARCO dataset, derived from Bing search queries, provides a large collection of real-world queries and their corresponding passages from relevant web documents. NQ contains real user queries from Google Search that are linked to relevant Wikipedia articles, emphasizing text retrieval for answering intricate information needs. For both these datasets, we follow the preprocessing approach of \cite{MINDER} and utilize pseudo queries generated from passages as docids for PIXAR. 

\textbf{Proprietary Dataset:} We further evaluate PIXAR in the context of sponsored search, where the objective is to retrieve relevant ads for user queries. We utilize advertiser bid keywords as docids for ads. We perform offline evaluations on SponsoredSearch-1B, a large-scale dataset of query-keyword pairs mined from the logs of a large commercial search engine. This dataset includes approximately 1.7 billion query-keyword pairs, with 70 million unique queries and 56 million unique keywords. The test set consists of 1 million queries, with a retrieval set of 1 billion keywords.

\textbf{Metrics \& Baselines:} Following prior work \cite{MINDER, LTRGR}, we evaluate all models using MRR@k and Recall@k for the MS MARCO dataset, and Hits@k for NQ. For the SponsoredSearch-1B dataset, we use Precision@K as the evaluation metric. Additionally, we measure inference latency with a batch size of 1 on a Nvidia T4 GPU. We compare PIXAR with several AR baselines, including DSI \cite{DSI}, NCI \cite{NCI}, SEAL \cite{SEAL}, MINDER \cite{MINDER}, and LTRGR \cite{LTRGR}. We report retrieval results from the respective papers and obtain inference latency by running the official code. For NAR baselines, we include CLOVERv2 \cite{unity} and replicate their method on our datasets due to the absence of reported numbers and official code for these datasets. Complete implementation details are provided in Appendix \ref{appendix}. 

\begin{table}[]
\centering
\def\arraystretch{1.1}%
\resizebox{0.99\textwidth}{!}{
\begin{tabular}{cccccccccc}
 \hline
\multirow{2}{*}{}    & \multirow{2}{*}{\textbf{Models}} & \multirow{2}{*}{\textbf{\begin{tabular}[c]{@{}c@{}}GPU\\ Latency\end{tabular}}} & \multicolumn{4}{c}{\textbf{MS MARCO}}                                                                                       & \multicolumn{3}{c}{\textbf{Natural Questions}}                                         \\ \cline{4-10} 
                     &                                  &                                                                                    & \multicolumn{1}{c}{@5}           & \multicolumn{1}{c}{@20}          & \multicolumn{1}{c}{@100}         & M@10        & \multicolumn{1}{c}{@5}            & \multicolumn{1}{c}{@20}           & @100          \\ \hline
\multirow{7}{*}{AR}  & DSI \cite{DSI, dsi_scaling}                             & -                                                                                  & \multicolumn{1}{c}{-}             & \multicolumn{1}{c}{-}             & \multicolumn{1}{c}{-}             & 17.3          & \multicolumn{1}{c}{28.3}          & \multicolumn{1}{c}{47.3}          & 65.5          \\  
                     & NCI \cite{NCI}                             & -                                                                                  & \multicolumn{1}{c}{-}             & \multicolumn{1}{c}{-}             & \multicolumn{1}{c}{-}             & 9.1           & \multicolumn{1}{c}{-}             & \multicolumn{1}{c}{-}             & -             \\  
                     & SEAL-LM \cite{SEAL}                         & 84.3x                                                                                & \multicolumn{1}{c}{-}              & \multicolumn{1}{c}{-}              & \multicolumn{1}{c}{-}              &  \multicolumn{1}{c}{-}              & \multicolumn{1}{c}{40.5}          & \multicolumn{1}{c}{60.2}          & 73.1          \\  
                     & SEAL-LM+FM \cite{SEAL}                      & 84.3x                                                                                & \multicolumn{1}{c}{-}              & \multicolumn{1}{c}{-}              & \multicolumn{1}{c}{-}              &    \multicolumn{1}{c}{-}            & \multicolumn{1}{c}{43.9}          & \multicolumn{1}{c}{65.8}          & 81.1          \\  
                     & SEAL \cite{SEAL}                            & 84.3x                                                                                & \multicolumn{1}{c}{19.8}          & \multicolumn{1}{c}{35.3}          & \multicolumn{1}{c}{57.2}          & 12.7          & \multicolumn{1}{c}{61.3}          & \multicolumn{1}{c}{76.2}          & 86.3          \\  
                     & MINDER \cite{MINDER}                          & 94.1x                                                                                & \multicolumn{1}{c}{29.5}          & \multicolumn{1}{c}{53.5}          & \multicolumn{1}{c}{78.7}          & 18.6          & \multicolumn{1}{c}{65.8}          & \multicolumn{1}{c}{78.3}          & 86.7          \\  
                     & LTRGR \cite{LTRGR}                           & 94.1x                                                                                & \multicolumn{1}{c}{\textbf{40.2}} & \multicolumn{1}{c}{\textbf{64.5}} & \multicolumn{1}{c}{\textbf{85.2}} & \textbf{25.5} & \multicolumn{1}{c}{\textbf{68.8}} & \multicolumn{1}{c}{\textbf{80.3}} & \textbf{87.1} \\  \hline
\multirow{3}{*}{NAR} & CLOVERv2 \cite{unity}                        & 1.0x                                                                                 & \multicolumn{1}{c}{29.2}          & \multicolumn{1}{c}{47.7}          & \multicolumn{1}{c}{66.9}          & 18.3          & \multicolumn{1}{c}{49.6}          & \multicolumn{1}{c}{63.4}          & 72.9          \\  
                     & PIXAR (Ours)                            & \textbf{1.2x}                                                                      & \multicolumn{1}{c}{\textbf{38.7}} & \multicolumn{1}{c}{\textbf{61.0}} & \multicolumn{1}{c}{\textbf{80.9}} & \textbf{24.0} & \multicolumn{1}{c}{\textbf{61.1}} & \multicolumn{1}{c}{\textbf{74.1}} & \textbf{82.7} \\  
                     & \% improvement                   & -                                                                                  & \multicolumn{1}{c}{32.7\%}        & \multicolumn{1}{c}{27.9\%}        & \multicolumn{1}{c}{20.9\%}        & 31.0\%        & \multicolumn{1}{c}{23.2\%}        & \multicolumn{1}{c}{16.9\%}        & 13.4\%        \\ \hline
\end{tabular}}

\caption{Performance and inference latency on MS MARCO and NQ. We report Recall@5, 20, 100, MRR@10 (MS MARCO) and Hits@5, 20, 100 (NQ), with inference latency relative to CLOVERv2. Bottom row shows PIXAR's relative improvement over CLOVERv2.  "-" denotes unreported results. \label{tab:main_results_msmarco_nq} 
}
\vspace{-0.8cm}
\end{table}
\subsection{Results}
We present the results of PIXAR and various GR baselines on the MS MARCO dataset in columns 4-7 of Table \ref{tab:main_results_msmarco_nq}. We observe several key findings from this comparison. First, CLOVERv2 significantly outperforms AR baselines like SEAL, NCI, and DSI, while also offering substantial improvements in inference latency. This highlights CLOVERv2 as a strong NAR baseline. However, CLOVERv2 falls short when compared to more recent AR models, particularly MINDER and LTRGR. For instance, CLOVERv2's recall at 100 is lower than that of MINDER by 11.8 absolute points. Next, our proposed PIXAR model with a 5M target vocabulary outperforms the strong CLOVERv2 baseline across all metrics, showing approximately 20-30\% relative improvements. This strongly supports our hypothesis that increasing the target vocabulary of NAR models can significantly imrpove retrieval performance. Moreover, PIXAR exceeds the performance of MINDER in every metric, achieving a 22.5\% improvement in MRR at 10, while also achieving substantial speedups in inference latency. Notably, PIXAR achieves this performance without utilizing multiple types of docids like MINDER (titles, n-grams, pseudo queries) and relies solely on pseudo queries. Additionally, PIXAR closely rivals LTRGR, lagging by only 1.5 absolute points in MRR@10 (a 5.8\% relative difference), despite not using a complex two-stage training with a passage-level loss like LTRGR.

The results on the NQ dataset are presented in the last three columns of Table \ref{tab:main_results_msmarco_nq}. Here, the baseline CLOVERv2 NAR model significantly trails behind AR models like SEAL, MINDER, and LTRGR. For example, CLOVERv2 exhibits a relative gap of 16.3\% with respect to LTRGR on recall at 100. Similar to MS MARCO, PIXAR substantially outperforms CLOVERv2 on all metrics, yielding around 13-23\% gains while maintaining significant latency speedups over AR models. Importantly, PIXAR reduces the relative gap with LTRGR from 16.3\% to 5.1\%. These results demonstrate the effectiveness of PIXAR in leveraging large vocabularies in NAR models to achieve substantially better retrieval performance than standard NAR models while retaining their latency benefits. 

\subsection{Ablations}
\label{subsec:ablation}
Our PIXAR model integrates three primary components: (i) a vocabulary and tokenizer that incorporate phrases in addition to words, (ii) an expanded vocabulary size of 5M tokens, and (iii) an efficient inference pipeline (Section \ref{subsection:pixar-pipeline}) to accelerate NAR inference. To analyze the impact of each component, we conducted detailed ablation studies, which we describe below.
\begin{wraptable}[]{}{0.5\textwidth}
\def\arraystretch{1.2}%
\resizebox{0.5\textwidth}{!}{
\begin{tabular}{l|c|c|c|c}
\hline
\textbf{Tokenizer}           & \textbf{M@10} & \textbf{R@5}  & \textbf{R@20} & \textbf{R@100} \\ \hline
DeBERTa       & 18.3          & 29.2          & 47.7          & 66.9           \\ 
BPE                 & 18.7          & 29.8          & 48.5          & 67.4           \\ 
Unigram             & 19.0          & 30.5          & 49.7          & 68.7           \\ \hline
Phrase-based & \textbf{21.6} & \textbf{34.7} & \textbf{56.0} & \textbf{77.5}  \\ \hline
\end{tabular}}
\caption{Retrieval performance of different tokenizers on MS MARCO (vocabulary size of 128K)}
\vspace{-0.12cm}
\label{tab:ablation_tokenizer}
\end{wraptable}
\begin{wraptable}[]{}{0.65\textwidth}
\centering
\def\arraystretch{1.1}%
\resizebox{0.65\textwidth}{!}{
\begin{tabular}{cccccccc}
\hline
\multirow{2}{*}{\begin{tabular}[c]{@{}c@{}}Vocab\\ Size\end{tabular}} & \multicolumn{4}{c}{\textbf{MS MARCO}}                                                                                                   & \multicolumn{3}{c}{\textbf{Natural Questions}}                                         \\ \cline{2-8} 
                                                                      & \multicolumn{1}{c}{@5}            & \multicolumn{1}{c}{@20}           & \multicolumn{1}{c}{@100}          & \multicolumn{1}{c}{M@10} & \multicolumn{1}{c}{@5}            & \multicolumn{1}{c}{@20}           & @100          \\ \hline
128K                                                                  & \multicolumn{1}{c}{34.7}          & \multicolumn{1}{c}{56.0}          & \multicolumn{1}{c}{77.5}          & 21.6                    & \multicolumn{1}{c}{56.7}          & \multicolumn{1}{c}{71.6}          & 80.7          \\ 
500K                                                                  & \multicolumn{1}{c}{34.9}        & \multicolumn{1}{c}{56.9}        & \multicolumn{1}{c}{78.6}        & 21.7                    & \multicolumn{1}{c}{57.8}          & \multicolumn{1}{c}{72.7}          & 81.4          \\ 
800K                                                                  & \multicolumn{1}{c}{35.2}        & \multicolumn{1}{c}{57.5}        & \multicolumn{1}{c}{79.2}        & 21.9                    & \multicolumn{1}{c}{58.2}          & \multicolumn{1}{c}{73.0}          & 81.2          \\ 
1M                                                                    & \multicolumn{1}{c}{35.7}          & \multicolumn{1}{c}{58.4}          & \multicolumn{1}{c}{79.6}          & 22.5                    & \multicolumn{1}{c}{58.5}          & \multicolumn{1}{c}{73.0}          & 82.0          \\ 
5M                                                                    & \multicolumn{1}{c}{\textbf{38.5}} & \multicolumn{1}{c}{\textbf{61.0}} & \multicolumn{1}{c}{\textbf{81.6}} & \textbf{24.2}           & \multicolumn{1}{c}{\textbf{61.2}} & \multicolumn{1}{c}{\textbf{74.8}} & \textbf{83.5} \\ \hline
\end{tabular}}
\caption{Scaling vocabulary improves NAR retrieval: We report the Recall@k and Hits@k for MSMARCO and NQ datasets}
\vspace{-0.15cm}
\label{tab:ablation_vocab_scaling}
\end{wraptable}

\textbf{Phrase-enhanced Vocabulary:} We first investigated the effectiveness of PIXAR's vocabulary construction strategy (detailed in Section \ref{subsec:vocab_selection}), focusing on the inclusion of phrases. To isolate this effect, we fixed the vocabulary size to 128K, equivalent to that of DeBERTa-v3, which was used to initialize the encoder. We compared the retrieval performance on the MS MARCO dataset using the original DeBERTa BPE tokenizer, a custom sub-word BPE, a sub-word Unigram, and our phrase-based tokenizer, all trained on the MS MARCO docid set. Table \ref{tab:ablation_tokenizer} presents the retrieval performance for the different tokenizers.

We observed that a custom-tailored BPE tokenizer performs marginally better than the original DeBERTa tokenizer. Further, the Unigram tokenizer outperforms the BPE by approximately 1.9\% in MRR@10 and Recall@100, in relative terms. Most notably, our phrase-based tokenizer substantially outperforms the best baseline (Unigram tokenizer), with a relative improvement of 13.7\% in MRR@10 (from 19.0\% to 21.6\%) and 12.6\% in Recall@100 (from 68.7\% to 77.5\%). These results clearly demonstrate the benefits of extending beyond words to include phrases in the vocabulary for NAR models. 

\textbf{Vocabulary Scaling:} Next, we analyze the impact of increasing the target vocabulary size in NAR models, addressing \textbf{RQ1} posed in Section \ref{section:intro}. For this study, we utilized the phrase-based tokenizer and varied the vocabulary size from 128K to 5 million tokens. We used the full softmax operation without any approximation to observe the raw effect of scaling. 
As shown in Table \ref{tab:ablation_vocab_scaling}, there is a consistent increase in retrieval performance as the vocabulary size increases across both MS MARCO and NQ datasets. Notably, the improvement persists even when the vocabulary size exceeds 1 million tokens. For instance, when increasing the vocabulary size from 1 million to 5 million tokens, Recall@5 on the MS MARCO dataset improves by 7.7\% (from 35.7 to 38.5). These findings highlight the clear advantages of scaling up the vocabulary size in NAR models in terms of retrieval performance.

\begin{wraptable}[]{}{0.55\textwidth}
\centering
\def\arraystretch{1.05}%
\resizebox{0.55\textwidth}{!}{
\begin{tabular}{ccccc}
\hline
\multirow{2}{*}{\textbf{Method}}  & \multicolumn{2}{c}{\textbf{MSMARCO}} & \multicolumn{2}{c}{\textbf{Latency (ms)}} \\ \cline{2-5}
  & MRR@10 & R@100 & Mean & 99 \\
\hline
Full Softmax & 24.2 & 81.6 & 47.9 & 48.3 \\
SVD-Softmax \cite{svdsoftmax} & 22.8 & 78.6 & 13.7 & 14.3 \\
HiRE-Softmax \cite{hire} & 24.0 & 81.3 & 12.7 & 13.2 \\
Centroid Clustering \cite{fvpc} & 21.7 & 78.2 & 14.2 & 17.4 \\ 
Fast Vocab  \cite{fvpc}  & 22.6 & 79.6 & 9.5 & 16.7\\
PIXAR (Ours) & 24.0 & 80.9 & 4.5 & 5.0 \\
\hline
\end{tabular}}
\label{table:softmaxopt}
\caption{Retrieval performance and inference latency (in ms) for various softmax optimization methods}
\label{tab:ablation_inference_optimizations}
\end{wraptable}
\textbf{Efficient PIXAR Inference:} Scaling vocabulary size introduces computational challenges due to the expensive softmax operation. Table \ref{tab:ablation_inference_optimizations} compares PIXAR's inference pipeline (Section \ref{subsection:pixar-pipeline}) against established techniques: (i) low-rank approximation methods: SVD-Softmax \cite{svdsoftmax}, HiRE-Softmax \cite{hire}) and (ii) clustering-based methods: Fast Vocabulary Projection  \cite{fastsoftmax} and it's variant Centroid Projection. While offering modest speedups, low-rank approximations like HiRE-softmax still result in significantly higher inference latency (3.4x slower than the 128k vocabulary CLOVERv2 baseline) due to their linear complexity with vocabulary size. Clustering-based methods like Fast Vocabulary Projection offer further speedups in mean latency but remain 2.5x slower than CLOVERv2. In contrast, PIXAR achieves superior performance, delivering a 10.6x speedup over full softmax and a 2.1x speedup over Fast Vocabulary Projection while maintaining comparable retrieval performance to full softmax (within 0.82\% in MRR@10 and 0.85\% in Recall@100). This translates to a latency only ~21\% higher than the CLOVERv2 model which has a ~39x smaller vocabulary. These results highlight the effectiveness of our tailored softmax approximation, which efficiently predicts shortlist tokens for all language modeling head projections in NAR models.

\subsection{Further Analysis}
To gain deeper insights into PIXAR's superior performance compared to smaller-vocabulary NAR models like CLOVERv2, we present qualitative examples in Table \ref{tab:examples}. PIXAR's tokenizer effectively captures multi-word entities like locations (e.g., "des moines iowa") and common phrases (e.g., "average temp", "what's the weather like in") as single tokens. Consequently, the weights in the language modelling head of PIXAR can learn representations for these multi-word entities and phrases from training data, capturing their semantic meaning. In contrast, standard NAR models like CLOVERv2 tend to break down words representing single concepts into multiple tokens (e.g., "des moines iowa" is fragmented into four tokens: "des", "mo", "ines", "iowa"). 
This hinders the language modeling head from learning meaningful representations for these concepts. Moreover, representing common phrases like "what's the weather like in" allows PIXAR to make fewer independent predictions in parallel, reducing the target output sequence length. Specifically, the mean and 99th percentile target sequence length decreases from 10.98 to 4.05 and from 18 to 9 in PIXAR compared to CLOVERv2. This reduction in target tokens simplifies the model's prediction task, leading to improved retrieval performance. Interestingly, despite shorter target sequence lengths, PIXAR tends to predict longer outputs with more words, as each token represents multiple words. This addresses a common issue with NAR models, namely their tendency to generate short outputs \cite{natshort1,unity}. More details on the sequence length and target sentence length analysis can be found in Appendix \ref{appendix}.
\begin{table}\
\centering
\def\arraystretch{1.3}%
\resizebox{1\textwidth}{!}{
\begin{tabular}{@{}P{3cm}|P{7.2cm}|P{7cm}@{}}
\hline
\setul{0.5ex}{0.3ex}
\textbf{Query} & \textbf{PIXAR} & \textbf{CLOVERv2} \\
\hline
average temperatures des moines iowa & 
 1. \setulcolor{red}\ul{average temp} \setulcolor{blue}\ul{des moines iowa} \newline
 2. \setulcolor{red}\ul{what's the average temperature in}  \setulcolor{blue}\ul{des moines iowa} \newline
3. \setulcolor{red}\ul{weather in} \setulcolor{blue}\ul{des moines iowa}  \setulcolor{blue}\ul{fahrenheit}\newline
 4. \setulcolor{red}\ul{what's the weather like in}  \setulcolor{blue}\ul{des moines}  &
 1. \setulcolor{red}\ul{average} \setulcolor{blue} \ul{temperature}  \newline
 2. \setulcolor{red}\ul{what}  \setulcolor{blue}\ul{temperature}     \newline
3. \setulcolor{red}\ul{what}  \setulcolor{blue}\ul{is}  \setulcolor{red}\ul{des}  \setulcolor{blue}\ul{mo}-\setulcolor{red}\ul{ines}  \setulcolor{blue}\ul{des} \setulcolor{red}\ul{mo}-\setulcolor{blue}\ul{ines} \newline
4. \setulcolor{red}\ul{what}  \setulcolor{blue}\ul{is}  \setulcolor{red}\ul{des} 
 \\
\hline
how many best - western points for free night & 
 1. \setulcolor{red}\ul{best western rewards points} \newline
 2. \setulcolor{red}\ul{how many best western} \setulcolor{blue}\ul{rewards points} \setulcolor{red}\ul{do i need} \newline
3. \setulcolor{red}\ul{how many best western} \setulcolor{blue}\ul{hotels} \newline
4. \setulcolor{red}\ul{how many best western} \setulcolor{blue}\ul{points} \setulcolor{red}\ul{for free nights} \newline & 
 1. \setulcolor{red}\ul{how} \setulcolor{blue}\ul{many} \setulcolor{red}\ul{hotels} \newline
 2. \setulcolor{red}\ul{what} \setulcolor{blue}\ul{is} \setulcolor{red}\ul{points}  \newline
 3. \setulcolor{red}\ul{how} \setulcolor{blue}\ul{many} \setulcolor{red}\ul{hotel}  \newline
 4. \setulcolor{red}\ul{how} \setulcolor{blue}\ul{many} \setulcolor{red}\ul{best} \setulcolor{blue}\ul{western} \setulcolor{red}\ul{points} \setulcolor{blue}\ul{for} \setulcolor{red}\ul{free} \setulcolor{blue}\ul{nights} \\
\hline
\end{tabular}}
\caption{Examples from PIXAR (5M vocab) and CLOVERv2 (128K vocab) on two sampled queries from MS MARCO dev set. Underlined spans indicate target tokenizer tokens. \label{tab:examples}}
\vspace{-0.4cm}
\end{table}

\subsection{Application to Sponsored search}
To demonstrate the effectiveness of PIXAR in real-world scenarios, we conducted a series of experiments in sponsored search, where the task is to retrieve the most relevant advertisements for user queries. In this application, ads are treated as documents, and the keywords bid by advertisers serve as the docids. We first evaluated PIXAR on the SponsoredSearch-1B dataset, where it significantly outperformed CLOVERv2, increasing P@100 from 23.5\% to 29.1\% (relative improvement of 23.7\%). Further, we deployed PIXAR on a large-scale commercial search engine and conducted A/B testing against an ensemble of leading proprietary dense retrieval and generative retrieval algorithms. PIXAR improved revenue by 4.02\% with 5.08\% increase in clicks, 0.64\% increase in click-through rate, and 4.35\% increase in query coverage, underscoring its effectiveness in a real-world setting. 
\section{Conclusion}

In this work, we introduced PIXAR, a novel NAR-based retrieval approach that leverages phrase-level tokens within an expanded target vocabulary. Our experiments demonstrated that PIXAR bridges the performance gap with state-of-the-art AR methods while retaining the inherent efficiency of NAR models. This speed advantage positions PIXAR as a promising candidate for latency-sensitive applications like real-time search and recommendation systems.

\bibliographystyle{acl_natbib}
\bibliography{references}

\newpage
\appendix
\label{appendix}
\section{Implementation Details}
\label{appendix:impl}

\begin{figure*}
    \includegraphics[width=0.7\textwidth]{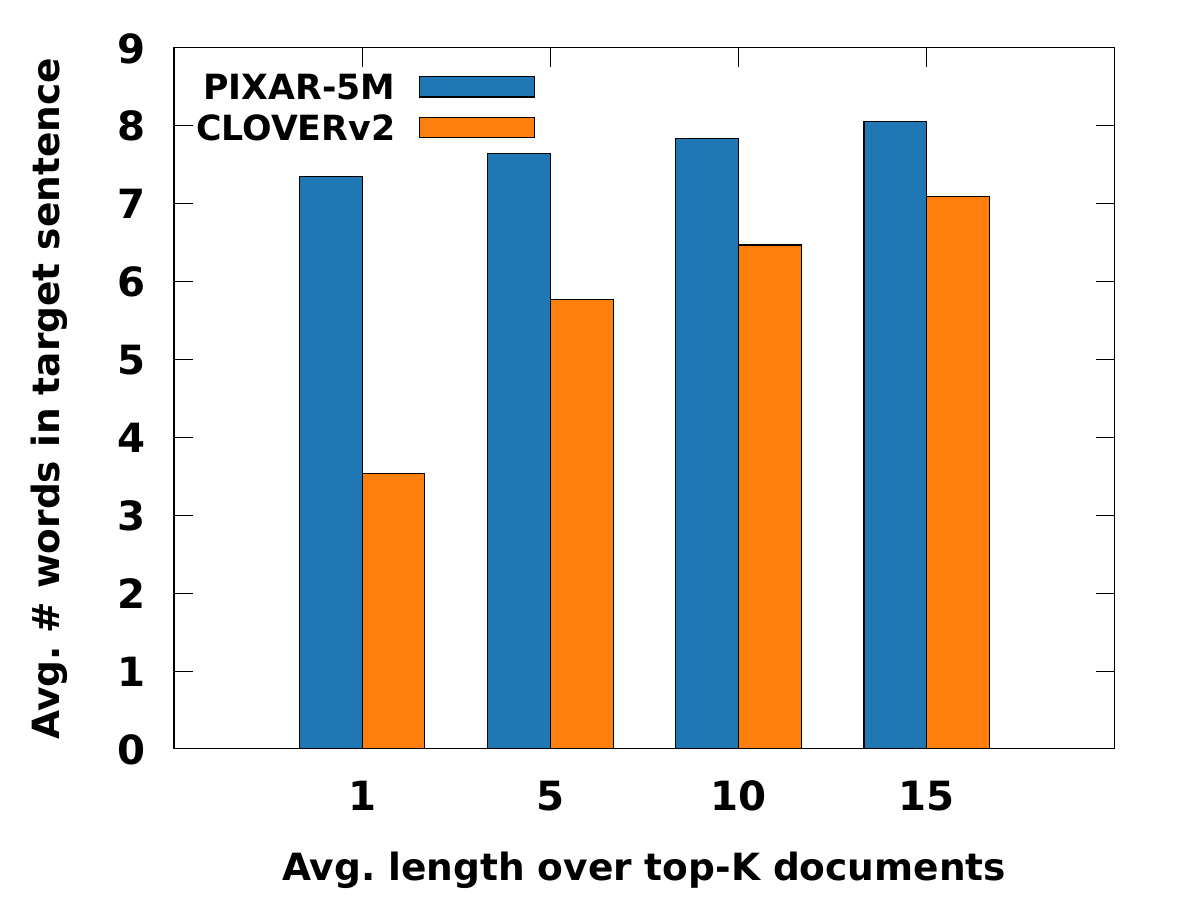}
    \caption{PIXAR-5M generates longer sentences on average compared to CLOVERv2 (NQ dataset)}
    \label{fig:longer-words}
\end{figure*}

\subsection{Model details}
We initialize PIXAR and the baseline CLOVERv2 model with the pretrained DeBERTa encoder \cite{deberta_v3}. We use the "microsoft/deberta-v3-base" checkpoint available on HuggingFace. For CLOVERv2, we use the provided 128K DeBERTa vocabulary for both the input and target. The language modeling head for PIXAR must necessarily be initialized from scratch.

\subsection{Document identifiers}
We employ the pseudo queries used in MINDER as our document identifiers. The total number of unique pseudo queries is around 80 million for the Natural Questions Wikipedia passages, and about 170 million for the MS MARCO passages. In addition to using pseudo queries as our document identifiers, we also augment our training dataset by adding these pseudo queries as questions that map to other pseudo queries asked of the same passage. For each passage, we sample up to 20 pseudo queries and add them to the training dataset. 

\subsection{Training details}
All models were trained with a learning rate of $5\times10^{-5}$, 1000 warmup steps, and an effective batch size of 6400. Hyperparameters $\lambda_2$ (shortlist loss scaling factor) and $\lambda_3$ (self-normalization loss scaling factor) were set to 0.25 and 1.0. The Adam optimizer was employed with a linear decay learning rate scheduler. Models were trained for 5 epochs on the MSMARCO dataset and 10 epochs on the Natural Questions dataset.
\subsection{Compute}
We trained models using a 5M target vocabulary on 8 Nvidia H100 GPUs and models of all other vocabulary sizes on 16 AMD Mi200 GPUs. Inference experiments were all carried out on an NVIDIA Tesla T4 GPU. Training time ranges from 1-2 days depending on the size of the vocabulary.

\subsection{Shortlisting module}
We set the hyperparameters $m, r, k$ to 4096, 20000 and 5 respectively.

\subsection{Vocabulary construction}
For PIXAR, we construct a target vocabulary of 5 million tokens using the method described in Section \ref{subsec:vocab_selection}. We construct separate vocabularies for MS MARCO and NQ datasets, on the full set of document identifiers for each dataset. TokenMonster binaries were used to construct the vocabulary. We detail some 
 important hyperparameters here. The "min-occur" parameter was set to 20 for constructing the PIXAR vocabulary, ensuring that candidate phrases occur at least 20 times in the document identifier corpus. While constructing the vocabulary, we use "strict" mode, in order to prevent minor variations of a phrase from receiving multiple tokens in the vocabulary.
\begin{wraptable}[]{}{0.5\textwidth}
\centering
\begin{tabular}{ccc}
\hline
\textbf{Model} & \textbf{Mean} & \textbf{99th} \\
\hline
\textbf{CLOVERv2} & 10.98 & 18 \\
\textbf{128K} & 5.56 & 12 \\
\textbf{500K} & 4.78 & 11  \\ 
\textbf{1M} & 4.46 & 10 \\
\textbf{5M} & 4.05 & 9  \\ 
\hline
\end{tabular}
\label{table:seqlen}
\vspace{1pt}
\caption{Mean and 99th percentile sequence lengths for different vocabulary sizes. The number of tokens needed to tokenize a document identifier nearly halves compared to the baseline.}
\label{merged_table}
\end{wraptable}

\subsection{Sequence Lengths and Target Sentence Lengths}
NAR models often generate document identifiers that are sometimes too brief to convey significant semantics. By contrast, PIXAR generates longer and more relevant target sentences, by generating phrases directly instead of subwords and words. Figure \ref{fig:longer-words} presents the aggregated results that shows that PIXAR (5M vocabulary). The phrase-based tokens in PIXAR have another benefit: they enable the generation of longer and relevant target sentences using fewer tokens, thereby enhancing generation quality. Table \ref{table:seqlen} illustrates  how the sequence lengths of the target tokens decrease as vocabulary sizes increase. Notably, the sequence lengths for the 128K vocabulary generated by PIXAR's vocabulary construction algorithm results in fewer token sequence lengths compared to CLOVERv2 which uses DeBERTa tokenization.
\newpage

\end{document}